%% file: 0-main.tex
\documentclass[11pt]{article}

\usepackage{amssymb}
\usepackage{lipsum}
\usepackage[table]{xcolor}
\usepackage{xcolor}
\usepackage{booktabs} 
\usepackage{tabularx}
\usepackage{amsmath}
\usepackage{float}
\usepackage{placeins}
\usepackage{url}
\usepackage[pdftex]{graphicx}
\usepackage{caption,subcaption}
\usepackage{hyperref}
\usepackage[numbers]{natbib}
\title{Finding Optimal Trading History in Reinforcement Learning for Stock Market Trading}

\author{Sina Montazeri\thanks{University of North Texas, Denton, Texas 76207, USA}
\and Haseebullah Jumakhan\thanks{Ajman University, Ajman, United Arab Emirates}
\and Amir Mirzaeinia\thanks{University of North Texas, Denton, Texas 76207, USA}}

\date{\today}

\begin{document}
\maketitle

\begin{abstract}
This paper investigates the optimization of temporal windows in Financial Deep Reinforcement Learning (DRL) models using 2D Convolutional Neural Networks (CNNs). We introduce a novel approach to treating the temporal field as a hyperparameter and examine its impact on model performance across various datasets and feature arrangements. We introduce a new hyperparameter for the CNN policy, proposing that this temporal field can and should be treated as a hyperparameter for these models. We examine the significance of this temporal field by iteratively expanding the window of observations presented to the CNN policy during the deep reinforcement learning process. Our iterative process involves progressively increasing the observation period from two weeks to twelve weeks, allowing us to examine the effects of different temporal windows on the model's performance. This window expansion is implemented in two settings. In one setting, we rearrange the features in the dataset to group them by company, allowing the model to have a full view of company data in its observation window and CNN kernel. In the second setting, we do not group the features by company, and features are arranged by category. Our study reveals that shorter temporal windows are most effective when no feature rearrangement to group per company is in effect. However, the model will utilize longer temporal windows and yield better performance once we introduce the feature rearrangement. To examine the consistency of our findings, we repeated our experiment on two datasets containing the same thirty companies from the Dow Jones Index but with different features in each dataset and consistently observed the above-mentioned patterns. The result is a trading model significantly outperforming global financial services firms such as the Global X Guru by the established Mirae Asset.
\end{abstract}

\tableofcontents


\input{1-Intro}

\input{2-Lit-Rev}

\input{2.1.hypothesis}

\input{3-Methodology}

\input{4-Eval}

\input{5-application}

\input{6-sum-con}





\section{Declaration of generative AI and AI-assisted technologies in the writing process}
During the preparation of this work the author(s) used ChatGPT  as writing assistant to draft text, improving clarity , proofreading, language refinement, and saving time. After using this tool/service, the author(s) reviewed and edited the content as needed and take(s) full responsibility for the content of the publication.



\bibliography{7-biblio}

\end{document}

%% file: 1-Intro.tex
\section{Introduction}

\subsection{Background and Motivation}
Integrating Deep Reinforcement Learning (DRL) in financial market analysis significantly evolved investment analysis with Deep Learning. DRL combines deep learning and reinforcement learning to offer a sophisticated framework for adapting strategies in the dynamic financial domain. It allows a deep learning model to effectively decipher complex patterns in historical market data often overlooked by traditional quantitative models.
It is no secret that financial markets are inherently complex and influenced by economic trends and geopolitical events. Therefore, traditional financial modeling often struggles to adapt to these ever-changing conditions. However, with its direct learning from data and adaptive strategies, DRL presents a promising solution to these challenges. With its autonomous learning ability and continual adaptation to the financial environment, it leverages historical market data to identify complex relationships and patterns.

\subsection{Overview of Our Previous Work}
In recent years, significant progress has been made in applying deep reinforcement learning (DRL) to stock trading strategies. For instance, Wang et al. proposed a parallel multi-module DRL algorithm that effectively captures both current market conditions and long-term dependencies using fully connected and LSTM layers \cite{parallel_drl_stock_trading}. Zhang et al. introduced an automated stock trading system based on the Proximal Policy Optimization algorithm, modeling trading as a Markov decision process \cite{novel_drl_stock_trading}. Additionally, Huang et al. demonstrated the importance of integrating market sentiment data to enhance the performance of DRL models in trading \cite{market_sentiment_drl_stock_trading}. Liu et al. developed an end-to-end trading strategy using a multi-view environment representation neural network, incorporating a Long Memory mechanism to improve decision-making \cite{drl_end_to_end_stock_trading}. Lastly, Li et al. focused on adaptive trading strategies using Gated Recurrent Units to capture time-series data effectively \cite{adaptive_drl_stock_trading}. These studies collectively highlight the potential of DRL in creating robust and adaptive trading strategies.

Liu et al. significantly advanced Deep Reinforcement Learning in Finance by developing platforms such as FinRL-Meta \cite{Liu2022FinRL}. This platform is a comprehensive tool for training and evaluating data-driven reinforcement learning agents within several simulated financial markets, offering a robust benchmarking system for algorithm comparison and facilitating the simulation of complex market conditions. The FinRL platform enables researchers to refine and test the efficacy of various DRL strategies, and it has been pivotal in democratizing access to sophisticated financial simulation tools and propelling research in financial analysis.

FinRL uses environments that offer broad simulation capabilities. These specialized environments, such as ABIDES-Gym \cite{Vyawahare2020}, provide the necessary infrastructure that allows FinRL to create discrete event simulations explicitly tailored for financial markets. ABIDES-Gym extends the OpenAI Gym interface to accommodate the complex dynamics of financial trading, allowing for a nuanced replication of market mechanisms and agent interactions. This level of detail will enable researchers and practitioners to explore the impact of individual agent behaviors and market responses, thus enhancing the understanding of market microstructure and agent-based modeling. The framework also streamlines the model training process on financial datasets, epitomizing the intersection of DRL and high-performance computing. It Leverages distributed computing resources to reduce training times significantly and optimizes computational workflows to enable the application of complex DRL models to extensive financial tasks. Their efforts have led to the creation of scalable and efficient financial models.

Our previous work \cite{Montazeri2023} demonstrated the efficiency and capability of CNNs when used as policies for deep reinforcement learning. We utilized the FinRL platform to conduct experiments on CNNs as a significantly improved policy to FinRL's original proposition. We also showed \cite{Montazeri2024, Montazeri2024GradientRC} that rearranging the stock market features used in the FinRL platform to group them per company could benefit the mode's performance. This study also utilizes the FinRL platform with its original dataset, containing features generated through traditional Technical Analysis used in Finance. It also uses the new dataset introduced in FinRL Meta, which contains statistically engineered features such as Simple Moving Average (SMA), momentum, and rate of change.

Building upon these foundational studies, our research aims to bridge the gap between CNN architecture optimization and financial market analysis. By introducing a systematic approach to temporal window selection, we seek to enhance the adaptability and performance of DRL models in capturing complex market dynamics.
    
\section{Objectives of the Current Study}
So far, we have presented the literature and the setting in which our study operates. The primary objective of our research is to explore the effects of changing the temporal window of a Convolutional Neural Network (CNN) used as a policy in a FinRL. By progressively expanding the observation period, beginning with a concise two-week window and incrementally enlarging it by two weeks in each iteration and culminating in twelve weeks, we aim to observe and analyze the performance of our model as its temporal window changes in the FinRL platform. This iterative window expansion is designed to examine the impact of different temporal scales on the model's performance. This process enables a comprehensive analysis of how varying lengths of financial data affect the model's predictive capabilities, offering insights and an opportunity to optimize the temporal granularity for financial market analysis. Our study also examines the arrangement of feature vectors within these expanding windows to better understand the model-market dynamics.

Furthermore, we contrast the model's performance across these different temporal windows to discern patterns in market behavior and model performance. In our study, short-term windows, particularly the initial two-week period, are hypothesized to be critical for understanding the model's ability to capture immediate market changes and short-term trends, which are essential for timely and accurate trading predictions. As the window expands, the model is expected to integrate a broader spectrum of market conditions, capturing longer-term trends and patterns. This bi-weekly expansion strategy is designed to balance the benefits of short-term immediacy and long-term historical perspective, ensuring the model remains adaptable and responsive to transient market fluctuations and enduring trends. We hope to contribute to financial analytics by demonstrating the efficacy of CNNs in a DRL setting and by providing new insights into the role of temporal dynamics in financial modeling.

%% file: 2-Lit-Rev.tex
\section{Literature Review}
\label{sec:litriv}

\subsection{Classic ML approachs}
When studying the progressive advancements in this field, the classical Machine Learning (ML) approach in financial analytics primarily revolves around statistical models that have formed the bedrock of quantitative finance. Linear regression, one of the most fundamental techniques, has been extensively utilized for predicting financial trends and stock prices. Its effectiveness in financial forecasting is documented in "Analysis of Financial Time Series" by Tsay \cite{Tsay2010}, offering a comprehensive understanding of linear models in finance. Moreover, decision trees have been widely employed for risk assessment and credit scoring, as demonstrated in the study by Kumar and Ravi \cite{Kumar2007}, showcasing their ability to handle categorical and continuous input variables effectively. However, despite their widespread application, these classical models often struggle with financial data's non-linearity and high dimensionality characteristic. This limitation, as highlighted in the survey by Atsalakis and Valavanis \cite{Atsalakis2009}, clearly indicates the need for more advanced approaches in capturing the complex dynamics of financial markets, especially in volatile or unpredictable scenarios.

\subsection{Neural Networks in DRL}
Integrating Neural Networks and Deep Reinforcement Learning (DRL) into financial market analysis represents a significant leap forward from traditional ML methods. As outlined in the groundbreaking work by Mnih et al. \cite{Mnih2015}, DRL combines the depth and complexity of deep neural networks with the decision-making prowess of reinforcement learning, creating a powerful tool for financial analysis. This approach, which allows for direct learning from vast amounts of unstructured market data, effectively identifying intricate patterns and trends, is a game-changer in the field. Convolutional Neural Networks (CNNs) application within DRL, in particular, has further advanced the field. CNNs, renowned for their ability to process high-dimensional sequential data, are highly effective in capturing temporal and spatial dependencies in financial time series. This is exemplified in the research by Tsantekidis et al. \cite{Tsantekidis2017}., which utilized CNNs to analyze and predict stock prices from limited order book data, demonstrating the model's proficiency in handling complex financial datasets. The success of DRL in financial applications lies in its ability to continually adapt and learn in an ever-changing environment, a crucial feazture given the dynamic nature of financial markets. 

Despite these advancements, there remains a gap in understanding how the temporal scope of input data affects CNN performance in financial DRL models. Our study addresses this gap by systematically exploring various temporal windows and feature arrangements.

%% file: 2.1.hypothesis.tex
Here is your hypothesis section with the references added:

```latex
\section{Hypothesis}

Convolution operations are fundamental to Convolutional Neural Networks (CNNs), which are particularly effective in processing data with a grid-like topology, such as images and sequential data \cite{courbariaux2016binarized} \cite{dai2021coatnet}. The convolution operation can be understood as a mathematical process that combines two sets of information. In the context of CNNs, this involves a convolutional kernel (or filter) moving across an input signal (such as an image or time series data) to produce a feature map.

Mathematically, for continuous signals, the convolution operation is defined as:

\[
(S * K)(t) = \int_{-\infty}^{\infty} S(\tau)K(t - \tau) d\tau
\]

Here, \(S\) represents the input signal, and \(K\) represents the convolutional kernel. This integral computes the area under the product of the two functions as the kernel slides over the input signal. However, in practical applications involving digital data, the signals are discrete, and thus the convolution operation is adapted to:

\[
(S * K)[n] = \sum_{m=-M}^{M} S[m]K[n - m]
\]

In this discrete form, the convolution operation involves summing the element-wise products of the input signal and the kernel as it moves across the input. The result is a new set of values (the feature map) that highlight certain features of the input signal, such as edges in an image or patterns in sequential data \cite{chen2019compressing}.

The size of the convolutional kernel (or filter) is a critical parameter in this operation. The kernel size determines the local region from which features are extracted. A larger kernel can capture more contextual information by encompassing a wider region of the input signal, while a smaller kernel focuses on finer details. The balance between capturing local and global features is essential for the performance of CNNs \cite{dai2016rfcn}.

Additionally, the padding applied to the input signal before convolution affects the output size and the nature of the features extracted. Padding involves adding extra values (typically zeros) around the input signal, which allows the kernel to process edge regions more effectively. The output size of the convolution operation is given by:

\[
O = \frac{N - K + 2P}{S} + 1
\]

where \(N\) is the input size, \(K\) is the kernel size, \(P\) is the padding, \(S\) is the stride (the step size of the kernel), and \(O\) is the output size. Properly setting these parameters ensures that the CNN can effectively learn and extract meaningful features from the input data \cite{chollet2017xception}. Understanding these concepts is crucial for optimizing CNN architectures, especially in settings where the observation window size can significantly impact the model's performance.

The performance of Convolutional Neural Networks (CNNs) in processing sequential data is significantly influenced by the size of the observation window used in the convolutional layers. The kernel size in a convolution layer determines the local region from which features are extracted. Larger kernels can incorporate more contextual information, but excessively large kernels may dilute distinct features. The optimization of window size can be expressed through the effective window size equation:

\[
W_{\text{eff}} = W_{\text{kernel}} + (W_{\text{kernel}} - 1) \times (D - 1)
\]

where \(W_{\text{eff}}\) is the effective window size, \(W_{\text{kernel}}\) is the kernel size, and \(D\) is the dilation factor.

Furthermore, the role of padding in convolution processes influences the spatial dimensions of the output feature map, described by:

\[
O = \frac{N - K + 2P}{S} + 1
\]

where \(N\) is the input size, \(K\) is the kernel size, \(P\) is the padding, \(S\) is the stride, and \(O\) is the output size. Excessive padding can lead to overemphasis on peripheral data and potential overfitting, similar to how an over-expanded window size may cause information overload, making distinct features less discernible:

\[
\text{Information Overload} \propto \frac{W_{\text{eff}}}{\text{Distinct Features}}
\]

Therefore, a crucial balance is needed between capturing local and global features. We hypothesize that the optimal selection of a temporal window size in a CNN balances local feature detection and global contextual understanding. An optimally sized window allows the model to effectively capture relevant features without succumbing to information overload or excessive generalization, thereby enhancing accuracy and performance in sequential data processing tasks \cite{chen2019compressing}.

Given that our CNN acts as a policy for a Deep Reinforcement Learning (DRL) algorithm, the window size as a hyperparameter will be optimized through reinforcement learning. This optimal window size is found at the point where local and global feature detection are balanced:

\[
\text{Optimal Window Size} \leftrightarrow \min \left( \Delta_{\text{Local-Global}} \right)
\]

where \( \Delta_{\text{Local-Global}} \) measures the differential in information capture between local and global features. This hypothesis suggests that through careful tuning and reinforcement learning, the CNN can achieve an optimal window size that maximizes performance in sequential data tasks.

%% file: 3-Methodology.tex
\section{Methodology}
In this study, we have integrated Deep Reinforcement Learning (DRL), Proximal Policy Optimization (PPO), and the Markov Decision Process (MDP) framework. The integration method is adopted from FinRL \cite{Liu2022FinRL}, providing a robust and dynamic model capable of navigating the complexities of financial markets. DRL offers the foundational learning mechanism, MDP provides a structured approach to decision-making in uncertain environments, and PPO ensures efficient and stable policy optimization. Together, these methodologies create a sophisticated model capable of learning, adapting, and optimizing trading strategies in real-time financial scenarios. The upcoming sections will describe each component in detail, beginning with an overview of DRL and its significance in our framework.

\subsection{Deep Reinforcement Learning (DRL)}
Deep Reinforcement Learning (DRL) integrates the pattern recognition capabilities of Deep Learning with the decision-making framework of Reinforcement Learning. This synergy enables the development of sophisticated models that can autonomously adapt to the complex and dynamic nature of financial markets, learning to optimize strategies based on data-driven insights. By leveraging vast and varied datasets, DRL models can identify latent patterns and trends, dynamically adjusting strategies by continually learning from market data. This ability to process high-dimensional data and make real-time decisions significantly advances over traditional quantitative approaches.

DRL's ability to respond to market volatility and changes is crucial in financial markets. It addresses the high dimensionality of financial data and the need for timely decision-making. This forms the basis for integrating Proximal Policy Optimization (PPO), which enhances the stability and efficiency of our learning process.

\subsection{Markov Decision Process (MDP)}
The Markov Decision Process (MDP) provides a mathematical framework for modeling decision-making in situations where outcomes are partly random and partly under the control of a decision-maker. MDPs are fundamental to understanding reinforcement learning and are particularly relevant in financial applications where decisions must be made under uncertainty. 

In our study, MDPs model the sequential decision-making process, where each action the agent takes affects future states and rewards. We represent the trading environment as an MDP with states, actions, and rewards meticulously defined to capture the intricacies of financial markets. The state space encapsulates key financial indicators, the action space comprises various trading actions, and the reward function reflects financial gains or losses. This representation allows our DRL model to effectively learn and optimize trading strategies over time, accounting for the probabilistic nature of financial markets and the impact of each decision on future market states.  With the MDP framework providing the foundation for decision-making, we now turn to the role of feature extraction in our DRL agent, specifically through Convolutional Neural Networks (CNNs).

\subsubsection{MDP Model for Stock Trading}
The trading market is a stochastic and interactive environment in nature and can be formulated as a Markov Decision Process (MDP) with state, action, and reward.
\begin{itemize}
    \item State $s=[b, p, h, f]:$ a set that consist of balance $b$, price $p \in \mathbb{R}_+^D$, holdings of stock $h\in \mathbb{Z}_+^D$, and fundamental indicators $f$. where D is the number of stocks that we consider in the market. Fundamental indicators covers financial ratios listed in tables \ref{table:sma_data}, \ref{table:tech_ind_data}.
    \item Action $a=[sell, buy, hold]:$ a set of actions for all D stocks, consisting of sell, buy, hold which leads to a reduction, growth, or no alteration in the holdings h, correspondingly.
    \item Reward $r(s, a, s'):$ The adjustment in portfolio value upon executing action "a" in state $"s"$ and transitioning to the next state $"s'"$. The portfolio value encompasses the total value of equities in the held stocks, denoted as $p^Th$, plus the remaining balance, $"b"$.
    \item Policy $\pi(s)$: The stock trading approach in state "s" entails the probability distribution of "a" in the state "s".
    \item The action-value function $Q_\pi(s, a)$
    represents the anticipated reward obtained by taking action "a" in state "s" according to policy $\pi$.
\end{itemize}
The primary objective of this process is to optimize (maximize) the reward. Various published approaches exist for addressing this challenge, each with its own set of advantages and disadvantages \cite{sutton2018reinforcement}. We select PPO  which is commonly used and show higher performance than other approaches.

\subsubsection{Proximal Policy Optimization (PPO)}
Proximal Policy Optimization (PPO) is a cornerstone of our methodology, providing a robust approach to policy gradient optimization. PPO iteratively updates the policy in a controlled manner, minimizing the cost function while ensuring minimal deviation from the previous policy. This approach is achieved through a clipped objective function, which restricts the extent of policy updates at each iteration. PPO maintains stability during the learning process by comparing the new policy's performance to the old policy and ensuring updates occur only if they improve performance within a specified margin. 

This stability is particularly crucial in the volatile context of financial markets, where significant, risky updates could destabilize the model. PPO's multiple epochs of stochastic gradient ascent optimizes the policy, enhancing sample efficiency by reusing data. This method is valuable in financial applications where data can be scarce and costly. Including entropy terms in the objective function encourages exploration, preventing premature convergence to suboptimal policies. This makes PPO an effective choice for our DRL framework, balancing exploration and exploitation to ensure consistent and reliable trading performance. With PPO demonstrated in our methodology, let us next discuss the role of the Markov Decision Process (MDP) in modeling decision-making under uncertainty.

\subsection{CNN is as a Feature Extraction Network}
The CNN integration into FinRL is facilitated through a specialized gym environment simulating stock trading scenarios. This environment includes quantitative elements of the stock market, such as stock prices, trading volumes, and various financial ratios, which are fed into the CNN for analysis, and the CNN's role within this environment is to extract high-level features from the input data, which are then utilized by the DRL agent to make trading decisions. By transforming raw financial data into meaningful features, the CNN enables the DRL agent to learn and optimize trading strategies effectively.

Within this framework, our previous work has demonstrated that using Convolutional Neural Networks (CNNs) in Deep Reinforcement Learning (DRL) for financial applications is notably effective. The CNN model processes input states comprising stock prices and technical indicators, capturing complex patterns and relationships within the data. This enables the model to autonomously learn and adapt strategies, making informed trading decisions based on a deeper understanding of market behavior.

The CNN acts as a feature extractor within the DRL framework. It processes the raw financial data, learning to identify relevant patterns and trends. These extracted features are then fed into the DRL agent, which uses them to make trading decisions. This integration allows the model to adapt its feature extraction process based on the rewards received, creating a dynamic learning system that evolves with changing market conditions.

\subsubsection{CNN Architecture}
As mentioned before, CNN architecture used in our study is designed to handle the multidimensional nature of financial data and train on extensive datasets. Leveraging convolutional layers, batch normalization, and ReLU activation functions enhances this model's feature extraction and pattern recognition robustness. CNN's ability to capture localized features and temporal dependencies is critical in financial markets rich in temporal dynamics and complex patterns.

The feature extraction process involves CNN identifying localized features and temporal dependencies within the financial data. Its ability to capture these dynamics ensures that the DRL agent can adapt its strategies in response to changing market conditions. The effectiveness of CNN as a feature extractor is further enhanced by its capacity to handle large datasets and complex input structures. This capability allows the model to leverage vast historical and real-time market data, improving its predictive accuracy and decision-making performance.

\begin{figure*}
    \includegraphics[width=1\columnwidth]{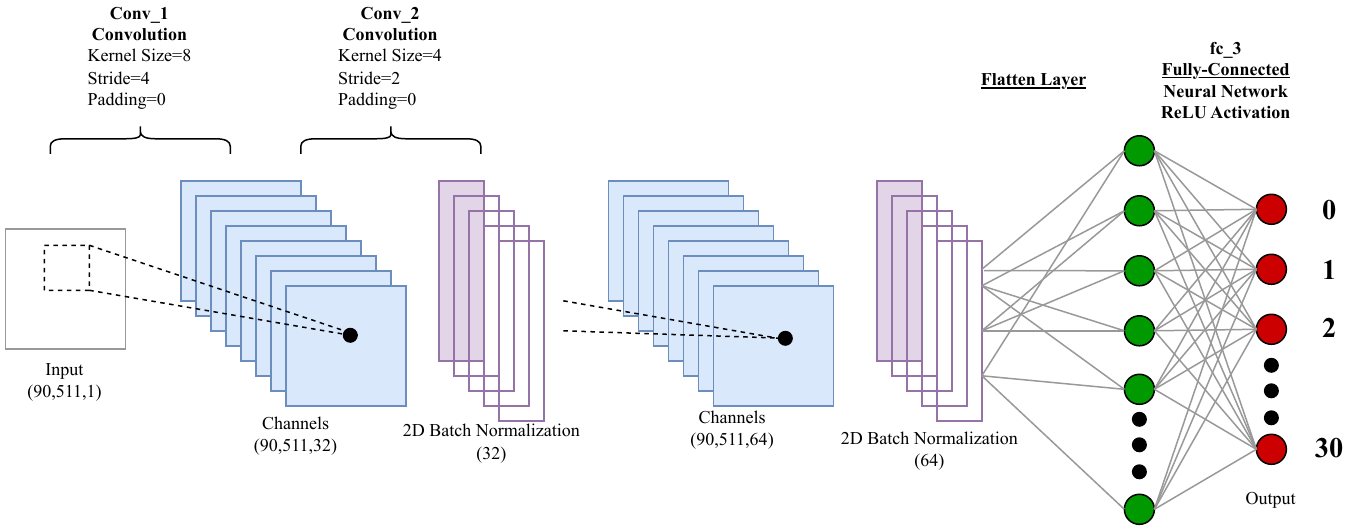}
    \caption{Architecture of the Convolutional Neural Network}
    \label{fig:architecture_of_network}
\end{figure*}

The network comprises two primary convolutional layers: the first layer features a kernel size of 8 and a stride of 4, while the second layer has a kernel size of 4 and a stride of 2. Both layers include 2D batch normalization, enhancing the network's efficiency in learning from the data by stabilizing the learning process. The data is then flattened and fed into a fully connected neural network layer with ReLU activation, integrating the extracted features for decision-making. The parameter specifications of our CNN network architecture are listed in Table \ref{tab:net_param}. 

The choice of this specific architecture was motivated by its ability to capture both short-term price movements and longer-term trends. The kernel sizes (8 and 4) were selected to allow the model to focus on weekly and monthly patterns, respectively, while the stride values (4 and 2) help in reducing computational complexity without significant loss of information.

\begin{table}[ht]
\centering
\rowcolors{2}{gray!10}{white}
{\fontsize{9pt}{11pt}
\selectfont
\begin{tabularx}{\linewidth}{Xll}
\toprule
\hiderowcolors
\textbf{Layer (type)} & \textbf{Output Shape} & \textbf{Parameter Size} \\
\showrowcolors
\midrule
Conv2d-1       & {[}-1, 32, 12, 85{]} & 2,080     \\
BatchNorm2d-2  & {[}-1, 32, 12, 85{]} & 64        \\
ReLU-3         & {[}-1, 32, 12, 85{]} & 0         \\
MaxPool2d-4    & {[}-1, 32, 6, 42{]}  & 0         \\
Conv2d-5       & {[}-1, 64, 12, 30{]} & 32,832    \\
BatchNorm2d-6  & {[}-1, 64, 12, 30{]} & 128       \\
ReLU-7         & {[}-1, 64, 12, 30{]} & 0         \\
MaxPool2d-8    & {[}-1, 64, 6, 15{]}  & 0         \\
Conv2d-9       & {[}-1, 128, 4, 13{]} & 73,856    \\
BatchNorm2d-10 & {[}-1, 128, 4, 13{]} & 256       \\
ReLU-11        & {[}-1, 128, 4, 13{]} & 0         \\
Conv2d-12      & {[}-1, 256, 2, 11{]} & 295,168   \\
BatchNorm2d-13 & {[}-1, 256, 2, 11{]} & 512       \\
ReLU-14        & {[}-1, 256, 2, 11{]} & 0         \\
Flatten-15     & {[}-1, 5632{]}       & 0         \\
Linear-16      & {[}-1, 1024{]}       & 5,768,192 \\
ReLU-17        & {[}-1, 1024{]}       & 0         \\
Dropout-18     & {[}-1, 1024{]}       & 0         \\
Linear-19      & {[}-1, 512{]}        & 524,800   \\
ReLU-20        & {[}-1, 512{]}        & 0         \\
Dropout-21     & {[}-1, 512{]}        & 0         \\
Linear-22      & {[}-1, 128{]}        & 65,664    \\
ReLU-23        & {[}-1, 128{]}        & 0         \\ \hline
\end{tabularx}%
}
\caption{Total model params: 6,763,552 \\
Trainable params: 6,763,552 \\
Non-trainable params: 0 \\
---------------------------------------\\
Input size (MB): 0.01 \\
Forward/backward pass size (MB): 1.74 \\
Params size (MB): 25.80 \\
Estimated Total Size (MB): 27.55 \\
---------------------------------------}
\label{tab:net_param}
\end{table}

With the CNN architecture established, the next step involves integrating this model into a DRL framework tailored for financial market analysis. This integration is facilitated by developing a specialized gym environment that simulates stock trading scenarios. The environment encapsulates critical elements of the stock market, including stock prices, trading volumes, and various financial ratios, which are fed into the CNN for analysis. This environment forms the backbone of our methodology, enabling the CNN to interact with and learn from a simulated financial market dynamically.

\subsection{Iterative Window Expansion Technique}

\begin{figure*}[htbp]
    \centering
    \includegraphics[width=\linewidth]{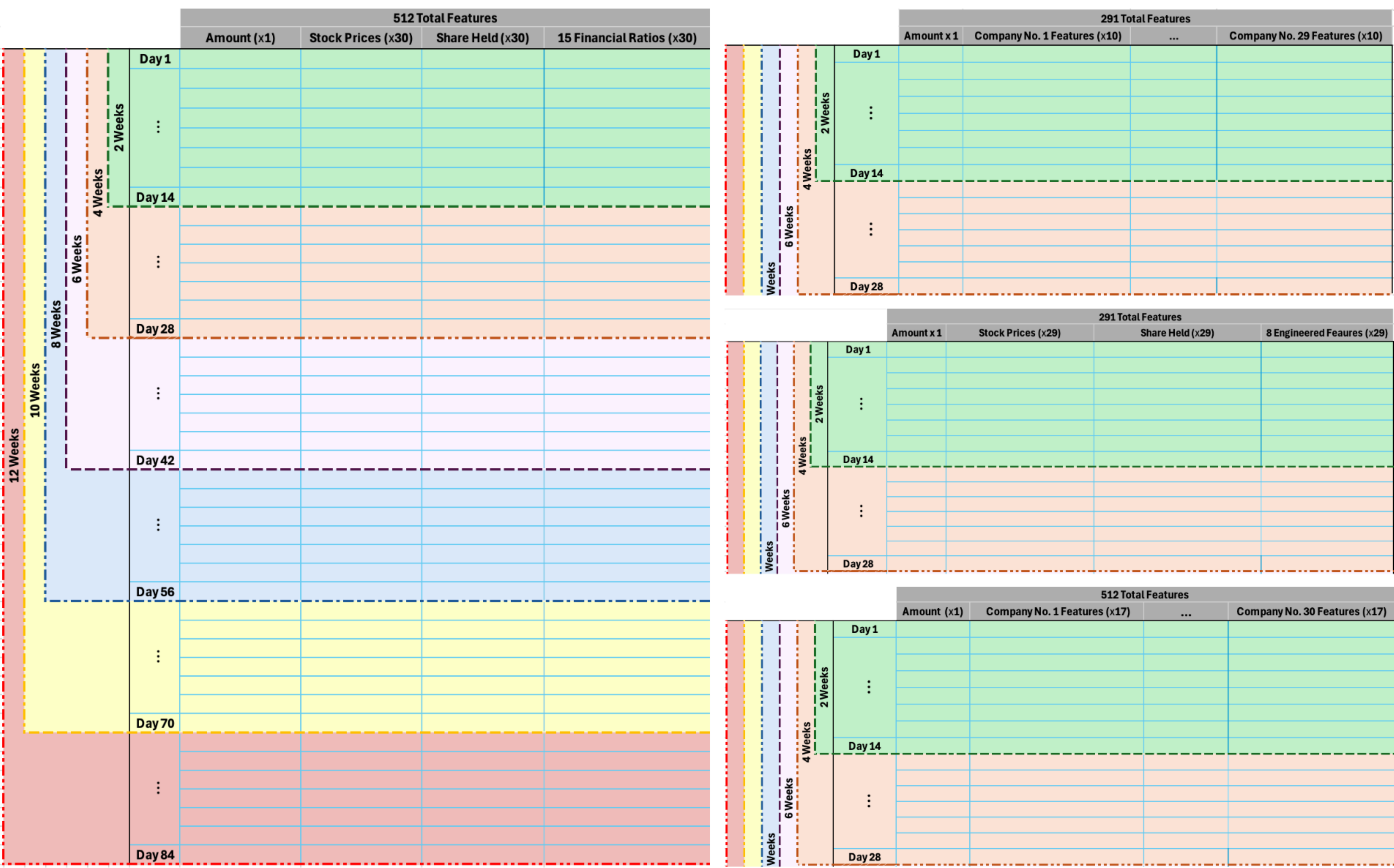}
    \caption{Composition of features, rearranged (left) and not rearranged (right)}
    \label{fig:journal_paper_features_old_data_not_rearranged}
\end{figure*}

We conducted 24 structured experiments across six temporal intervals ranging from 2 to 12 weeks, in 2-week increments. Each interval was chosen in 2-week increments, providing a range of short- to medium-term observations. We utilized two distinct dataset types for each timeframe: the Technical Indicator dataset and the Simple Moving Average (SMA) dataset. While these datasets encompass the same companies and timeframes, they include different features for each company. Each dataset was analyzed under two scenarios: one with rearranged features, grouping all columns associated with a single company, and another without rearrangement. This dual-path strategy, uniformly applied across all intervals, resulted in 24 unique experimental setups, comprehensively evaluating the CNN's performance and robustness under various temporal and data scenarios (Figure \ref{fig:journal_paper_features_old_data_not_rearranged}).

\subsubsection{Initial Two-Week Window}

The study begins with a concise two-week observation window, targeting short-term market trends to establish a baseline for model performance. This initial phase is critical for understanding the model's responsiveness to recent market changes and its ability to capture short-term patterns. The two-week window helps the model make timely and accurate predictions in the fast-paced financial trading environment by focusing on the most recent and relevant data points.

\subsubsection{Bi-Weekly Expansion Strategy}

Following the initial phase, we bi-weekly expanded the observation window, incrementally integrating more historical data. This gradual enlargement enables the model to assimilate information from a widening scope of market conditions, capturing more extensive long-term trends and patterns. The bi-weekly increments strike a careful balance, incorporating fresh data while retaining the benefits of an extended historical view. This approach ensures the model remains agile, effectively responding to immediate market changes and more substantial, enduring trends.

\subsubsection{Final Twelve-Week Window}

The process culminates with a twelve-week window, providing an exhaustive perspective on market trends and behaviors over an extended duration. This elongated observation period supplies the model with a diverse and comprehensive dataset, reflecting a broad spectrum of market activities. The value of the twelve-week window lies in its ability to reveal longer-term market trends and cyclical patterns, which are pivotal for strategic decision-making in financial trading. This concluding phase is crucial for evaluating the model's capacity to generalize and maintain consistent performance across various market cycles.

\subsection{Rearranged Features Approach}

Expanding upon our previous research, this paper also investigates the impact of feature rearrangement within expanded temporal windows. The rearranged features setting entails reorganizing the columns of the input data tensor to keep related features in proximity to each other. This arrangement aims to boost the capability of the CNN in identifying relevant patterns from the data, aligning with the underlying relationships and correlations inherent in financial indicators. Presenting the CNN with inputs specifically structured to reflect the interconnected nature of financial metrics is expected to enhance the model's accuracy and generalization ability. This preprocessing strategy is particularly pertinent in financial data analysis, where the interactions between various data types (such as stock prices, transaction volumes, and technical indicators) are often more critical than the individual data points. This approach is intended to promote more efficient learning, improving the model's robustness and adaptability when deployed on a wide range of financial datasets.

\subsection{Our Datasets}

To ensure the robustness of our approach, we utilized two distinct datasets from the FinRL and FinRL Meta projects. This methodology helps confirm that the success of our methods is not merely coincidental.

\subsubsection{The SMA Dataset}

The first SMA data dataset is adapted from the FinRL Meta project. This dataset encompasses quantitative financial features, including fundamental market data such as opening, high, low, and closing prices and trading volume. Additionally, it includes a series of engineered features like MACD, Bollinger Bands, RSI, CCI, and DX over 30 days, in addition to the 30-day and 60-day closing simple moving averages (SMAs), the VIX, and a turbulence measure. This rich compilation provides an extensive perspective on market trends and volatility, crucial for the Convolutional Neural Network (CNN) model's analysis across varying timeframes.

\begin{table}[ht]
\centering
\rowcolors{2}{gray!10}{white}
{\fontsize{9pt}{11pt}\selectfont
\begin{tabularx}{\columnwidth}{Xl}
\toprule
\hiderowcolors
\textbf{Dataset Column} & \textbf{Column Description} \\
\showrowcolors
\midrule
tic & Ticker symbol \\
open & Opening price \\
high & Highest price \\
low & Lowest price \\
close & Closing price \\
volume & Trading volume \\
day & Day of the week \\
macd & MACD value \\
boll\_ub & Upper Bollinger Band \\
boll\_lb & Lower Bollinger Band \\
rsi\_30 & 30-day Relative Strength Index \\
cci\_30 & 30-day Commodity Channel Index \\
dx\_30 & 30-day Directional Movement Index \\
close\_30\_sma & 30-day closing Simple Moving Average \\
close\_60\_sma & 60-day closing Simple Moving Average \\
vix & VIX value \\
turbulence & Market turbulence \\
\bottomrule
\end{tabularx}
}
\caption{Simple Moving Average (SMA) Data}
\label{table:sma_data}
\end{table}

\subsubsection{Feature Vector: A Trading Day in the Market}
Each trading day in the stock market includes a feature vector comprising the initial monetary amount, stock prices of twenty-nine companies, their corresponding shares held, and a set of eight quantitative features for each company. These indicators include MACD, Bollinger Bands (upper and lower), RSI 30, CCI 30, DX 30, 30-day and 60-day SMAs. The total feature vector comprises 261 elements: one for the initial amount, 29 for stock prices, 29 for shares held, and 232 derived from technical indicators (eight per company). Integrating these technical indicators, which play a critical role in signaling market trends and momentum, equips the dataset as an essential tool for the CNN model. It enables the model to identify and leverage market trends effectively, facilitating precise predictions and informed decision-making in the dynamic financial trading environment.

\begin{table}[!ht]
\centering
\rowcolors{2}{gray!10}{white}
{\fontsize{9pt}{11pt}\selectfont
\begin{tabularx}{\columnwidth}{Xl}
\toprule
\hiderowcolors
\textbf{Feature Name}                            & \textbf{Size} \\
\showrowcolors
\midrule
Amount                                   & 1             \\
Price                                    & 29            \\
Share held                               & 29            \\
MACD                                     & 29            \\
Bollinger Upper Band (boll\_ub)          & 29            \\
Bollinger Lower Band (boll\_lb)          & 29            \\
RSI 30                                   & 29            \\
CCI 30                                   & 29            \\
DX 30                                    & 29            \\
Close 30 SMA                             & 29            \\
Close 60 SMA                             & 29            \\
\text{Total size of feature vector}      & 261           \\
\bottomrule
\end{tabularx}
}
\caption{Daily Feature Vector for SMA Data}
\end{table}
\FloatBarrier

\subsubsection{The Technical Indicator Dataset}

The Technical Indicator Dataset offers an in-depth view of financial performance metrics, distinguishing itself from the SMA dataset with a more extensive set of financial ratios and metrics. While it also includes fundamental trading data such as opening price, highest price, lowest price, closing prices, and trading volume, its uniqueness lies in incorporating a diverse range of financial ratios. These include Operating and Net Profit Margins, Return on Assets, Return on Equity, various liquidity ratios (Current, Quick, Cash), turnover ratios (Inventory, Accounts Receivable, Accounts Payable), Debt Ratio, Debt to Equity Ratio, and market valuation ratios like PE, PB, and Dividend Yield. This dataset is instrumental in offering a detailed assessment of instruments' financial health and market valuation, a critical aspect of the nuanced market analysis conducted by our Convolutional Neural Network (CNN) model.

\begin{table}[!ht]
\rowcolors{2}{gray!10}{white}
{\fontsize{9pt}{11pt}\selectfont
\begin{tabularx}{\columnwidth}{Xl}
    \toprule
    \hiderowcolors
    \textbf{Dataset Column} & \textbf{Description} \\
    \showrowcolors
    \midrule
    tic & Ticker symbol \\
    open & Opening price \\
    high & Highest price \\
    low & Lowest price \\
    close & Closing price \\
    volume & Trading volume \\
    OPM & Operating Profit Margin \\
    NPM & Net Profit Margin \\
    ROA & Return on Assets \\
    ROE & Return on Equity \\
    cur\_ratio & Current Ratio \\
    quick\_ratio & Quick Ratio \\
    cash\_ratio & Cash Ratio \\
    inv\_turnover & Inventory Turnover \\
    acc\_rec\_turnover & Accounts Receivable Turnover \\
    acc\_pay\_turnover & Accounts Payable Turnover \\
    debt\_ratio & Debt Ratio \\
    debt\_to\_equity & Debt to Equity Ratio \\
    PE & Price to Earnings Ratio \\
    PB & Price to Book Ratio \\
    Div\_yield & Dividend Yield \\
    \bottomrule
\end{tabularx}
}
\caption{Technical Indicator Data Columns Description}
\label{table:tech_ind_data}
\end{table}
\FloatBarrier

\subsubsection{A Different Feature Vector}
The daily feature vector within this dataset is structured to provide an exhaustive market perspective through a multidimensional data array. This table comprises several components: the initial amount, stock prices of thirty companies, the number of shares currently owned in the simulation, and fifteen distinct financial ratios for each of the thirty companies. These ratios, extracted from each company's financial statements, offer vital insights into their financial performance. The feature vector, encompassing the data for one trading day in the stock market, contains 511 elements: one for the initial amount, 30 for stock prices, 30 for shares held, and 450 derived from the financial ratios (15 per company). This elaborate dataset is essential for the CNN model, enabling the analysis and interpretation of intricate patterns and correlations within the financial markets.

\begin{table}[ht]
\rowcolors{2}{gray!10}{white}
{\fontsize{9pt}{11pt}\selectfont
\begin{tabularx}{\columnwidth}{Xc}
\toprule
\hiderowcolors
\textbf{Feature Category} & \textbf{Number of Features} \\
\showrowcolors
\midrule
Amount                          & 1    \\
Price                           & 30   \\
Share held                      & 30   \\
Financial ratios (15 * 30)      & 450  \\
Total size of feature vector    & 511  \\
\bottomrule
\end{tabularx}
}
\caption{Technical Dataset Feature Vector}
\end{table}
\FloatBarrier

%% file: 4-Eval.tex
\input{results_table}
\FloatBarrier
\section{Results}
\subsection{Experimentation on the Technical Indicator Dataset}
The analysis of the Technical Indicator dataset, without any feature rearrangement, as illustrated in the figure below, uncovers a notable pattern in the accumulation of rewards over different time intervals. The most significant gain, observed in the 2-week observation size, reached a cumulative reward of 155.89. This finding highlights the efficacy of this specific observation window. The peak performance noted within this 2-week timeframe may constitute the most advantageous period for analysis in the context of this dataset and its feature composition. This observation window provides the optimal balance mentioned in our hypothesis section, generating the most significant rewards in the given feature arrangement setting and dataset.

\begin{figure}[ht]
\centering
\includegraphics[width=\linewidth]{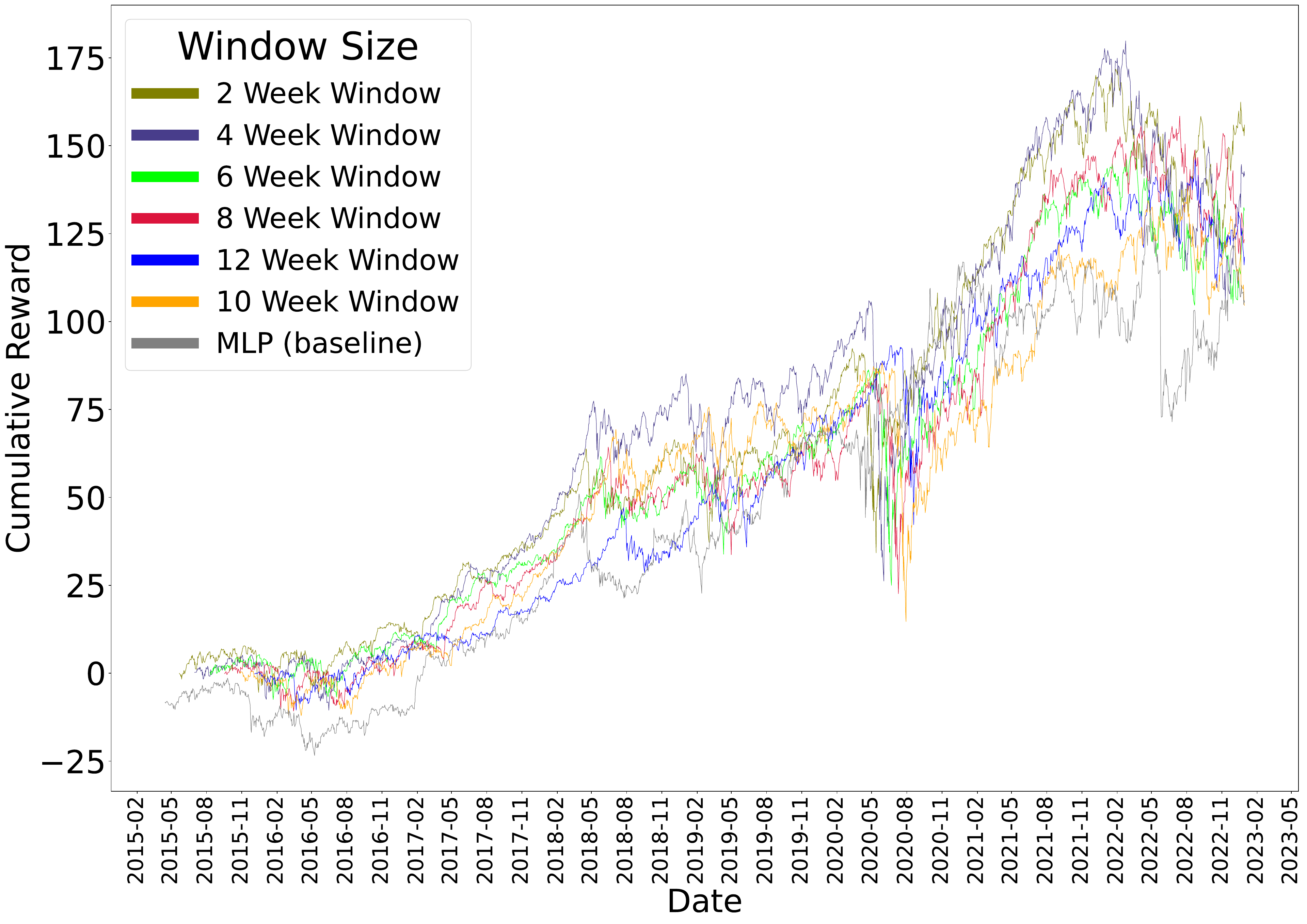}
\caption{Cumulative rewards in the Technical Indicator dataset without rearrangement }
\label{fig:tech_indicator_not_rearranged}
\end{figure}

The extended analysis of the Technical Indicator dataset over periods ranging from 4 to 12 weeks reveals a discernible decline in cumulative rewards, reaching its lowest point at the 10-week interval, where the reward significantly drops to 104.58. This downward trajectory, although slightly mitigated in the 12-week observation window, predominantly suggests diminishing returns as the duration of the observation period increases. This pattern serves as a crucial insight, highlighting the limitations of the convolutional neural network (CNN) in effectively utilizing longer observation windows for this specific dataset and feature configuration. This trend underscores the importance of strategically selecting the observation window to optimize the CNN's predictive performance, and it supports our hypothesis that information overload can diminish the CNN's ability to utilize most important features in the input tensor.

During the analysis of the Technical Indicator dataset with rearranged features, as depicted in the figure below, we found a markedly different trend in cumulative rewards across varying timeframes compared to the dataset with the original feature arrangement. The rearranged dataset demonstrates a similar pattern, where the peak cumulative reward is noted at the 10-week mark, registering at 121.59. This outcome indicates that the rearrangement of features shifts the optimal observation window to bigger sizes. Notably, a prolonged 10-week period emerges as most favorable in the rearranged dataset, in stark contrast to the 2-week window size identified as optimal in the original dataset configuration. This finding suggests that feature rearrangement significantly improves the model's ability to utilize longer observation windows, again underscoring the need for adaptable strategies in financial data analysis with CNNs.

\begin{figure}[ht]
    \centering
    \includegraphics[width=\linewidth]{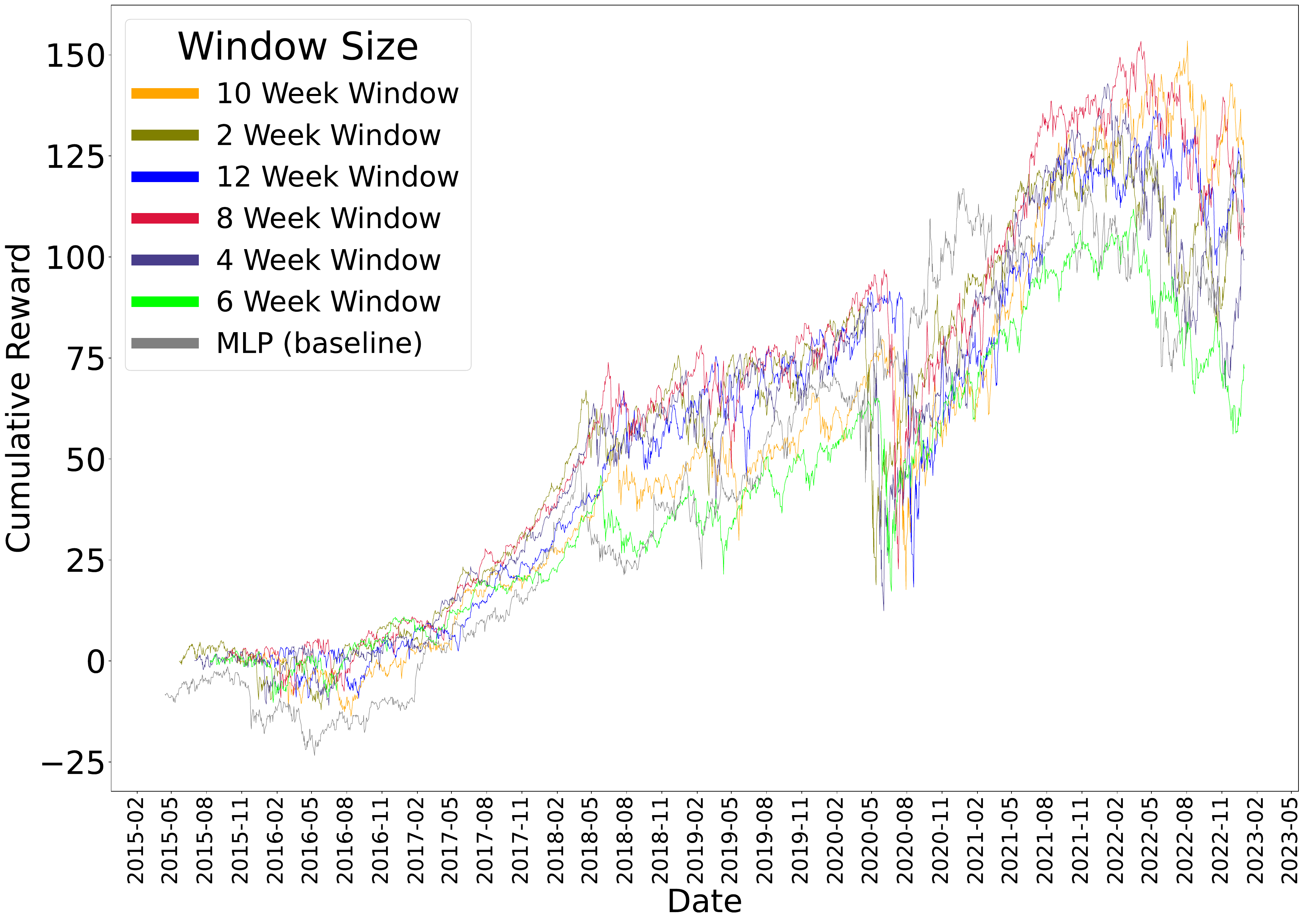}
    \caption{Cumulative rewards in the Technical Indicator dataset with rearrangement }
    \label{fig:tech_indicator_rearranged}
\end{figure}

As depicted in the figure, rearranging features within the technical indicator dataset markedly improves the model's capacity to capitalize on extended observation windows. Notably, the model's optimal performance, demonstrated at the 10-week interval with a cumulative reward of 121.59, signifies an enhanced ability to utilize more extended periods for analysis. This reorganization of features enables a more efficient interpretation of extended-term trends, optimizing the model's accuracy over such durations. This finding emphasizes the vital importance of feature engineering in amplifying the effectiveness of Convolutional Neural Networks, particularly in intricate and dynamic settings like financial market analysis.

In contrast, a different pattern emerges when analyzing the technical indicator dataset without feature rearrangement, as illustrated in the corresponding plot. Here, the 2-week interval emerges as the most productive timeframe, registering the highest cumulative reward of 155.89. This finding indicates that in its original configuration, the dataset is optimally tuned for short-term analysis, showing diminishing performance with lengthening observation periods, except for a slight increase at 12 weeks. However, these extended periods do not outperform the initial 2-week observation window. This trend highlights the model's predisposition towards shorter timeframes when processing the non-rearranged data, underscoring the impact of data structuring on the model's temporal adaptability and predictive power.

\begin{figure}[ht]
    \centering
    \includegraphics[width=\linewidth]{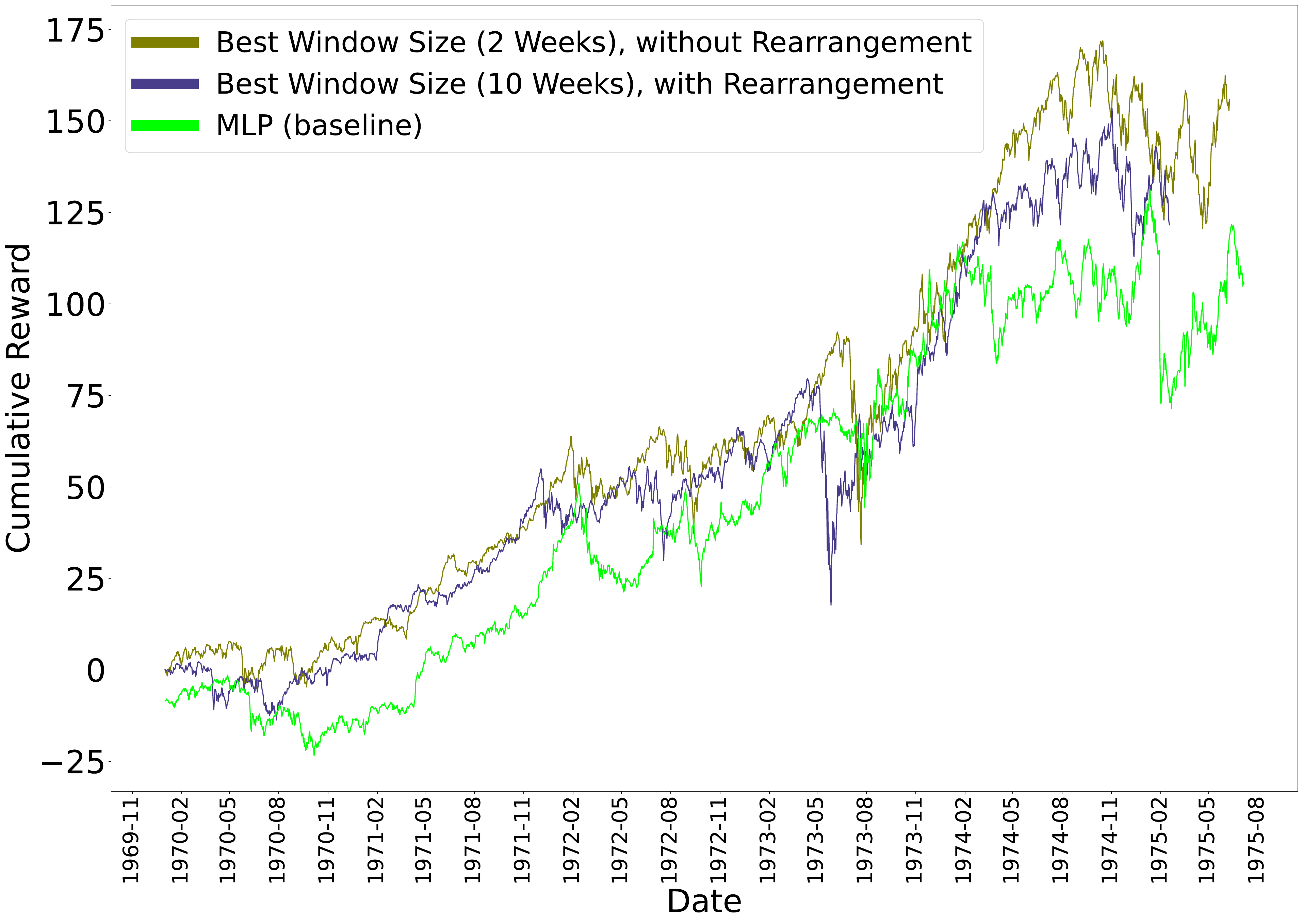}
    \caption{Best performers in the Technical Indicator Dataset}
    \label{fig:sma_nonrearranged}
\end{figure}

The contrasting results observed in the rearranged technical indicator data are striking. In this scenario, the model strides in the 10-week observation period, achieving a cumulative reward of 121.59. This shift from the optimal 2-week period in the non-rearranged data to a more extended 10-week period in the rearranged data is significant. The rearranging of features profoundly influences the model's efficiency in capturing and forecasting market trends. Compared to the reduced effectiveness in shorter durations, the enhanced performance at this longer interval underscores the impact of data sequencing on the model's predictive precision. This observation again stresses the criticality of data arrangement and preprocessing in financial time series analysis, as it can substantially alter the model's interpretation and response to market dynamics over different temporal scales. 

\subsection{Experimentation on the SMA dataset}
\begin{figure}[ht]
    \centering
    \includegraphics[width=\linewidth]{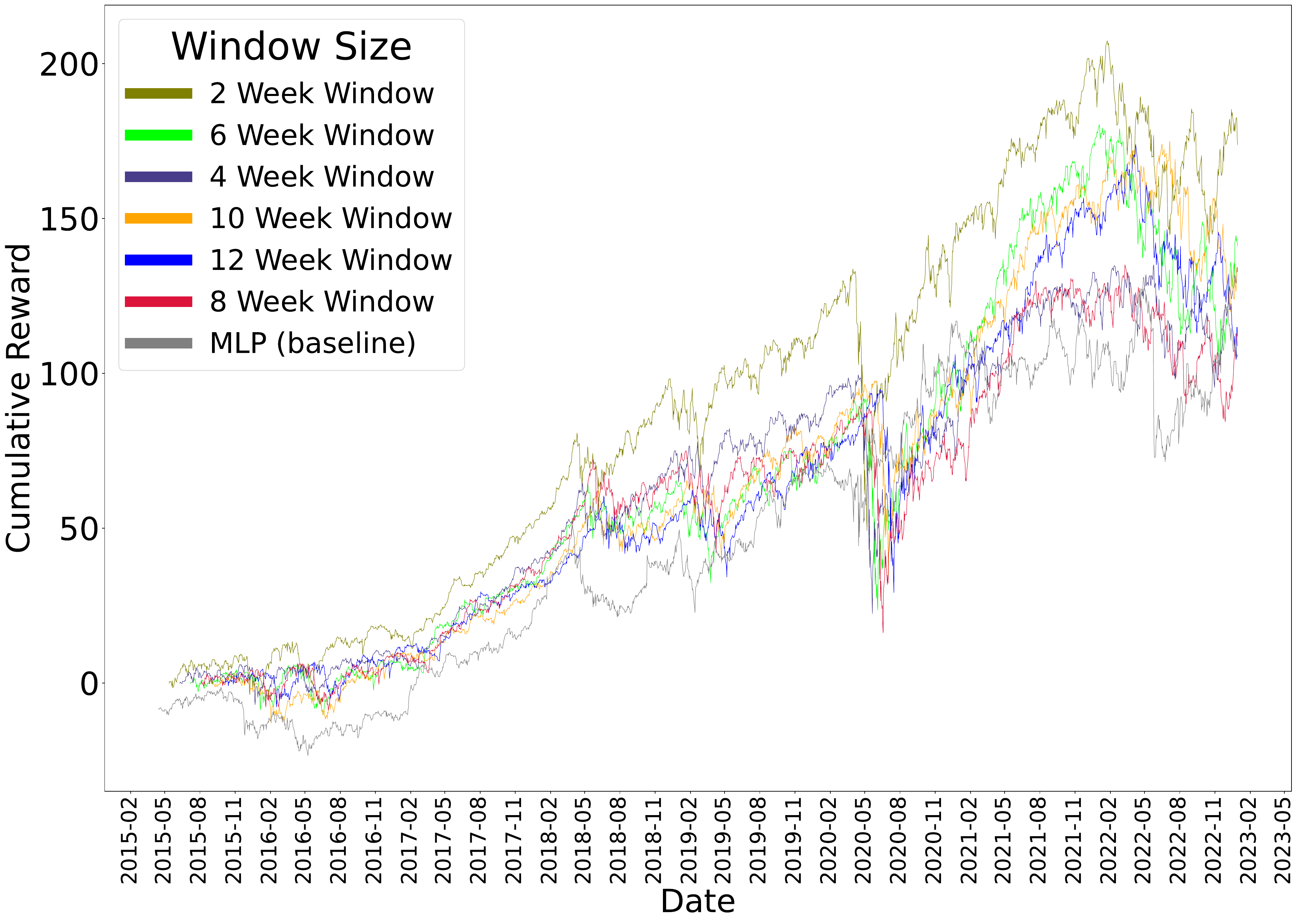}
    \caption{Cumulative rewards in the SMA dataset without rearrangement}
    \label{fig:sma_rearranged}
\end{figure}

The analysis of the SMA dataset without data rearrangement reveals a distinct pattern in cumulative rewards over various timeframes, as shown in Figure \ref{fig:sma_rearranged}. The most significant performance is apparent in the 2-week observation window, achieving a peak cumulative reward of 184.05. This high point suggests that a 2-week observation window is particularly effective for this dataset, indicating an optimal short-term period for analysis in this context.

As the observation window extends, a decreasing trend in cumulative rewards is evident, particularly at 8 and 12 weeks, with rewards noted at 99.80 and 105.99, respectively. However, an unexpected increase in cumulative reward to 144.22 at the 10-week mark presents an intriguing anomaly. This inconsistency might indicate complex, possibly cyclical patterns in the SMA dataset, which the model discerns differently across various intervals. This behavior further highlights the intricate nature of these quantitative indicators and emphasizes the importance of selecting an appropriate observation window for predictive modeling.

A different outcome is observed in the analysis of the SMA dataset with feature rearrangement. The 4-week interval emerges as the most favorable, registering a peak cumulative reward of 181.84. This result contrasts the lower performance in the 2-week window, where the cumulative reward is 117.14. This discrepancy suggests that rearranging the data may significantly alter the model's ability to utilize temporal relationships in the data, affecting its effectiveness across different timeframes. The rearranged dataset's peak at a longer interval underlines the same pattern where feature arrangement enhances the model's ability to effectively capture and analyze market trends.

\begin{figure}[ht]
    \centering
    \includegraphics[width=\linewidth]{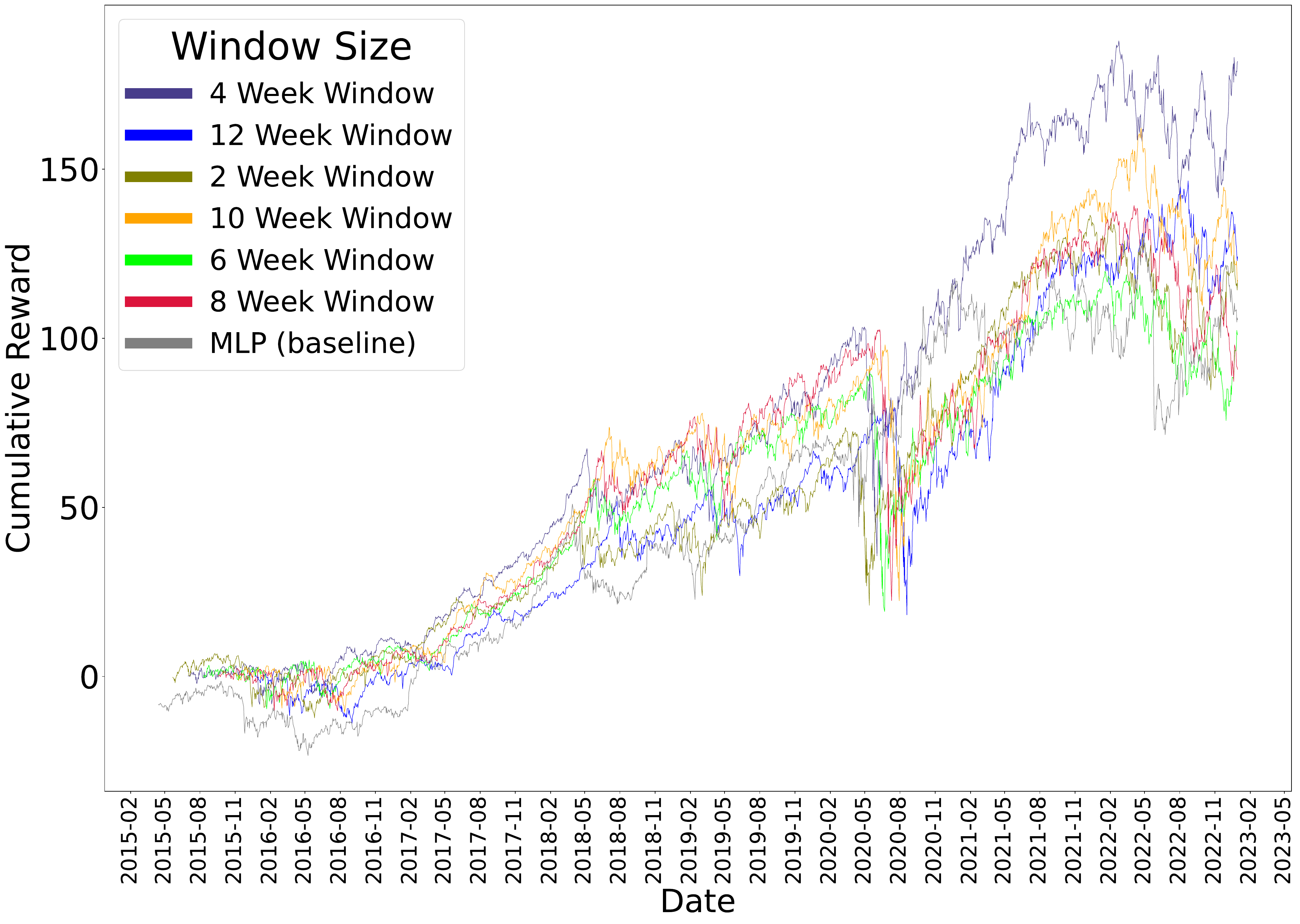}
    \caption{Cumulative rewards in the SMA dataset with rearrangement }
    \label{fig:new_data_no_suhffle}
\end{figure}

However, an irregular trend emerges as the observation period extends beyond 4 weeks. A marked decrease in cumulative rewards is noted at 6 and 8 weeks, with figures falling to 101.04 and 90.77, respectively. Intriguingly, there is a modest reward recovery at the 10 and 12-week intervals. This pattern suggests that the model may interpret different characteristics of the rearranged SMA dataset over extended timeframes. Such fluctuations in performance underscore the added complexity due to data rearrangement and the importance of carefully choosing the observation window to maximize the model's efficacy. 

\subsection{Best Performers in the SMA Dataset}
In the next phase of our data analysis, we conducted a comparative study of optimal timeframes in the simple moving average (SMA) dataset, considering its original and rearranged forms, as shown in the plot. This revealed distinctive trends. 

In the case of the non-rearranged SMA dataset, the most effective timeframe emerges as the 2-week window, registering a peak cumulative reward of 184.05. This notable performance at the shorter interval indicates the model's ability to effectively capture the prevailing trends within the original SMA data structure. As the observation period extends, a gradual decline in cumulative rewards is observed across longer timeframes. Although there is a marginal uplift in performance at the 10-week mark, this is within the benchmark set by the 2-week observation window, which means that the pattern still highlights the dataset's responsiveness to short-term fluctuations.

The clear differentiation in performance across various timeframes suggests that the underlying dynamics of the SMA dataset are more readily discernible and exploitable in shorter intervals when the data remains in its original sequence. This insight is pivotal for financial analysts and modelers, emphasizing the need for strategic consideration of time windows in predictive modeling, especially when dealing with complex financial datasets like the SMA.
\begin{figure}[ht]
    \centering
    \includegraphics[width=\linewidth]{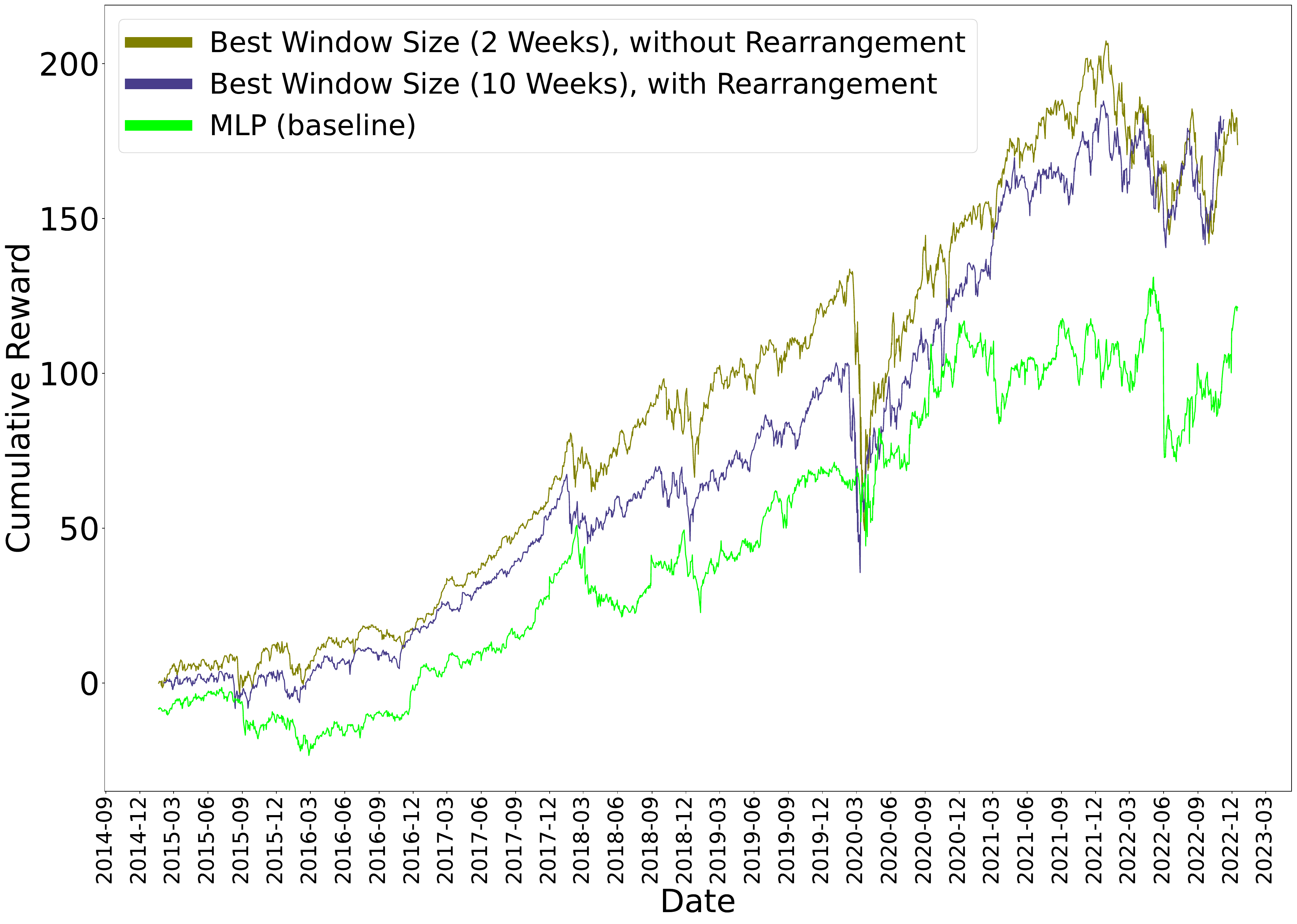}
    \caption{Best performers in the SMA dataset }
    \label{fig:sma_data_best_performers}
\end{figure}

In contrast, after rearranging the features in the SMA dataset, our analysis presents a different optimal timeframe. The 4-week window emerges as the best performer with a cumulative reward of 181.84, indicating a significant shift in the model's ability to utilize longer temporal windows. Our analysis also shows a more pronounced decline in performance for other timeframes, especially at 6 and 8 weeks. We noted that the 2-week observation size was the best performer in the non-rearranged data versus the 4-week peak in the rearranged data. Once again, the sharp contrast between the non-rearranged and rearranged data demonstrates the model's temporal processing ability.

\subsection{Best Performers overall}
Several insightful trends emerge in our final analysis of the datasets, encompassing both the SMA and Technical Indicator datasets. In its original feature arrangement, the SMA dataset exhibits strong performance in the 2-week timeframe, reaching a cumulative reward of 184.057, the highest across all datasets and timeframes. This result underscores the effectiveness of short-term observation in capturing market dynamics with this dataset. On the other hand, when the SMA features are rearranged, the 4-week window becomes the most productive, achieving a cumulative reward of 181.84. This shift suggests that market dynamics are captured more effectively over shorter temporal windows, but once features are rearranged, a slightly extended observation size proved more effective.

\begin{figure}[ht]
    \centering
    \includegraphics[width=\linewidth]{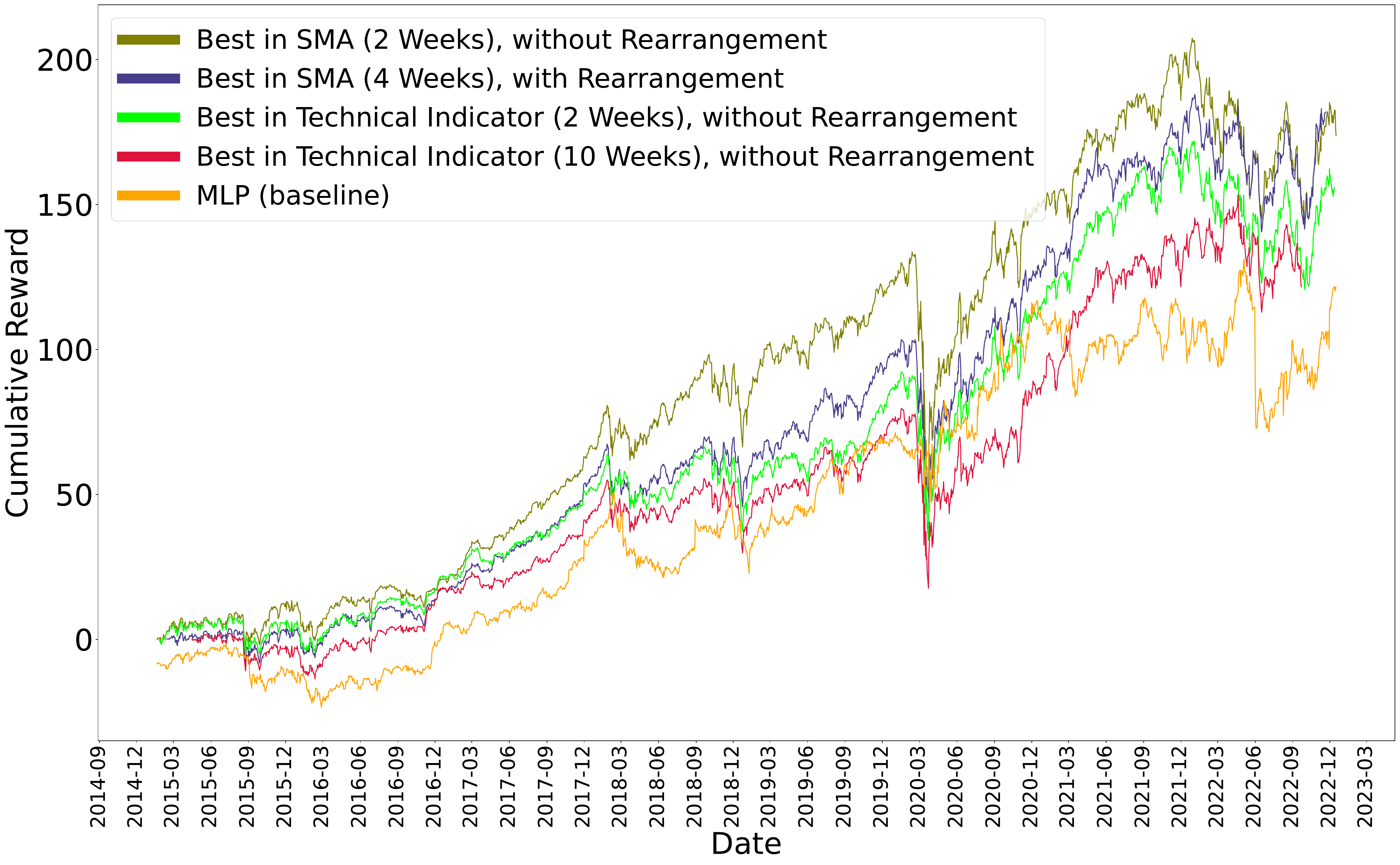}
    \caption{Best performers overall. }
    \label{fig:best_performers_overall}
\end{figure}

The observed trends in the Technical Indicator dataset echo those seen in the SMA dataset, particularly in the context of the original sequence. A 2-week observation window demonstrates optimal effectiveness, reaching a peak cumulative reward of 155.89. This similarity across the datasets consistently proves our decerned pattern that, without shorter observation periods, can be highly effective for predictive modeling. However, a significant shift occurs when the data sequence in the Technical Indicator dataset is rearranged. This modification leads to the 10-week timeframe becoming the most favorable, as evidenced by a cumulative reward of 121.59.

This shift indicates that the Convolutional Neural Network (CNN) becomes more adept at discerning the complex patterns between features and their temporal dynamics when the data is organized to maintain a cohesive structure for each company's features. The rearrangement enhances the model's ability to grasp longer-term trends and relationships, which may be less apparent or accessible in shorter timeframes or with non-rearranged data. This observation is crucial as it suggests that the efficacy of a CNN in financial market analysis can be significantly influenced by how the data is structured. It highlights the importance of considering the arrangement of data to optimize the predictive capabilities of models, especially in financial contexts where the relationships between various indicators and their evolution over time are crucial to understanding market movements. Thus, a flexible and context-specific approach to selecting observation periods and organizing data is paramount to maximizing the utility and accuracy of predictive models in financial analysis.

%% file: results_table.tex
\begin{table}[htbp]
\resizebox{\columnwidth}{!}{
\begin{tabular}{|l|cc|cc|}                              \hline

         & \multicolumn{2}{c|}{SMA}                                                 & \multicolumn{2}{c|}{Technical Indicator}                                 \\ \hline
         
         & \multicolumn{1}{l|}{with}                 & \multicolumn{1}{l|}{without} & \multicolumn{1}{l|}{with}                 & \multicolumn{1}{l|}{without} \\ \hline

2 weeks  & \multicolumn{1}{c|}{117.1}                & { \textbf{173.8}}         & \multicolumn{1}{c|}{120.6}                & { \textbf{155.9}}         \\ \hline

4 weeks  & \multicolumn{1}{c|}{{ \textbf{181.8}}} & 134.1                        & \multicolumn{1}{c|}{99.3}                 & 141.5                        \\ \hline
6 weeks  & \multicolumn{1}{c|}{101}                  & 141.1                        & \multicolumn{1}{c|}{72.5}                 & 131.6                        \\ \hline
8 weeks  & \multicolumn{1}{c|}{90.8}                 & 108.9                        & \multicolumn{1}{c|}{107.5}                & 123.3                        \\ \hline
10 weeks & \multicolumn{1}{c|}{116.6}                & 132.9                        & \multicolumn{1}{c|}{{ \textbf{121.6}}} & 104.6                        \\ \hline
12 weeks & \multicolumn{1}{c|}{124.1}                & 112                          & \multicolumn{1}{c|}{112.1}                & 118.3                        \\ \hline
\end{tabular}%
}
\caption{Cumulative Rewards in SMA and Technical Indicator Datasets for with and without Rearrangement}
\label{tab:cumulative_Rewards}
\end{table}

%% file: 5-application.tex
\section{Application}

\subsection{The Current Landscape of Hedge Funds and the Challenge of GURU ETF}
 Hedge funds have been pivotal in financial markets, known for their sophisticated strategies and adaptability. They employ a variety of tactics like taking long or short positions, relying on a thorough analysis of market trends, sector dynamics, company fundamentals, macroeconomic factors, and investor sentiment, supported by quantitative models and risk management.

Meanwhile, exchange-traded funds (ETFs) like the Global X Guru ETF (GURU) attempt to mirror the strategies of top hedge funds. GURU aims to replicate the stock picks of these funds based on their quarterly filings. Despite the allure of tapping into successful hedge fund strategies, GURU has struggled to match the performance of broader indices such as the S\&P 500, largely due to the delays in reporting and the inability to adjust to market changes in real-time.

\subsection{CNN-DRL to the Rescue}
Our CNN-DRL model presents a compelling alternative. The model excels in processing high-dimensional sequential data and adapts to various temporal windows, notably shorter ones, which is crucial in the fast-paced financial markets. The application of the CNN-DRL model within hedge funds could revolutionize their investment decision-making process, enhancing GURU's cost efficiency and performance.

The below chart illustrates the performance divergence between GURU, the S\&P 500, the CNN-DRL Model's best performer, and the DIA ETF. The CNN-DRL Model's best performer shows heightened growth compared to the steady rises of the S\&P 500 and DIA ETF, indicating a robust return on investment. Notably, it also demonstrates resilience during market downturns, avoiding the deep troughs experienced by GURU.

\begin{figure}[ht]
    \centering
    \includegraphics[width=\linewidth]{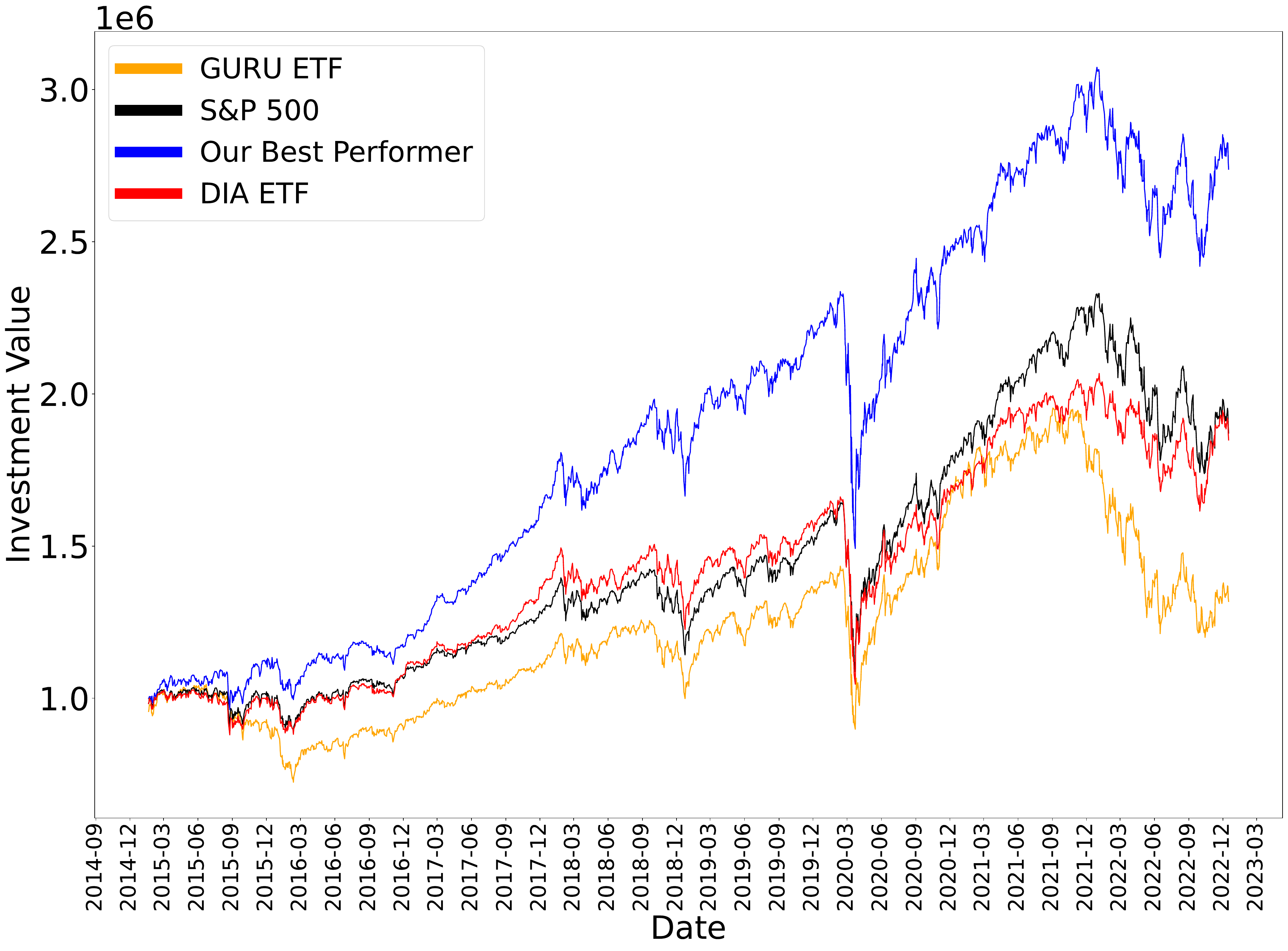}
    \caption{Comparison with traditional ETFs}
    \label{fig:comparison_with_dia}
\end{figure}

The CNN-DRL Model could offer a transformative edge to hedge fund strategies. Integrating this advanced system could allow funds to capture subtle market movements and respond with greater agility, resulting in higher returns and potentially lowering risk profiles. As financial landscapes grow more complex, adopting sophisticated models like the CNN-DRL may become essential for maintaining a competitive advantage.

For the Global X Guru ETF, the CNN-DRL Model could have significantly improved its strategy. Traditional methods, which may be less adaptive, could be enhanced by the CNN-DRL Model's capability to continuously learn and adapt, potentially leading to better investment decisions and growth in investment value.

The chart, depicting a comparison with traditional ETFs, serves as a visual testament to the potential enhancements that the CNN-DRL Model could bring to the investment strategies of ETFs like GURU and even the broader market-representative DIA ETF.

The superior performance of our CNN-DRL model, particularly in shorter temporal windows, has significant implications for high-frequency trading strategies. Fund managers could potentially use this approach to make more nimble, data-driven decisions in rapidly changing market conditions. Moreover, the model's ability to adapt to different feature arrangements suggests it could be applied across various financial instruments and markets, offering a versatile tool for portfolio management.

%% file: 6-sum-con.tex
\section{Discussion}

\subsection{Interpretation of Results}
The outcomes of this study, utilizing Convolutional Neural Networks (CNNs) within a Deep Reinforcement Learning (DRL) framework for financial analytics, underscore the pivotal role of temporal precision in market predictions. The findings particularly emphasize the efficacy of short-term observation windows, with the two-week window demonstrating superior performance in capturing market dynamics. This observation resonates with the rapidity and volatility characteristic of financial markets, where new information is swiftly reflected in stock prices. Notably, the enhanced model performance achieved through feature rearrangement highlights the significant impact of feature engineering. By reorganizing features related to different stocks and technical indicators, we support the hypothesis that a more robust and generalizable representation of data can be learned, potentially increasing the model's adaptability to diverse market conditions.

\subsection{Theoretical Implications}
The success of the CNN model, employing a methodical window expansion technique, accentuates the importance of temporal dynamics in financial time-series analysis. This approach aligns with the efficient market hypothesis, positing that markets assimilate all available information into stock prices. The model's ability to adjust to the market's 'memory' is crucial for precise financial forecasting. Additionally, the effectiveness of the rearranged features approach suggests that the organization and representation of information are as critical as the information itself for the learning process. This sheds light on the interaction of features within CNN layers, stressing the role of feature engineering in refining financial models.

Our findings challenge the conventional wisdom that longer observation windows invariably lead to better predictions in financial markets. The superior performance of shorter windows, particularly in non-rearranged datasets, suggests that recent market information may carry more predictive power than extended historical data. This aligns with the concept of market efficiency, where new information is rapidly incorporated into prices.

\section{Conclusions and Future Work}

In this study, we explored the impact of different data structures and observation windows on Convolutional Neural Networks (CNNs) performance in financial market analysis. We focused on understanding how the arrangement of data and the selection of timeframes influence the model's ability to capture and predict market dynamics.

The shift in optimal performance with rearranged data suggests that a Convolutional Neural Network (CNN) becomes more proficient at identifying complex patterns between features and their temporal dynamics when the data maintains a cohesive structure for each company's features. The ability of the model to grasp longer-term trends and relationships in rearranged datasets, which are less apparent in shorter timeframes or non-rearranged data, emphasizes the importance of strategic data arrangement. This finding underlines the need for flexible and specific approaches in selecting observation periods and organizing data to enhance the utility and accuracy of CNNs in financial market analysis.

\subsection{Limitations and Challenges}

This study, while offering valuable insights into the use of Convolutional Neural Networks (CNNs) for financial market analysis, encounters several limitations and challenges that warrant attention. A primary constraint is the need for increased computational power. Exploring additional observation windows and testing larger, more complex CNN architectures necessitate substantial computational resources. This requirement becomes particularly critical when considering the intricacy and volume of financial data and the need for extensive testing to validate the robustness and accuracy of the models across various market scenarios.

Moreover, there is a significant need for research funding to support these endeavors. Enhanced funding would facilitate access to more powerful computing infrastructure and enable a broader scope of experimentation. This includes investigating a wider array of temporal windows and deploying more advanced CNN models, which could potentially uncover deeper insights and yield more precise predictive capabilities.

Future studies could explore incorporating diverse data types, like news sentiment or economic indicators, to enhance model robustness. Broadening the scope to different markets and asset types would help verify the applicability of these findings. Investigating the effects of shorter temporal windows or real-time data streams may provide insights into high-frequency trading strategies. Additionally, applying the concept of rearranged features to other forms of financial data, such as order book information or unstructured data, could pave the way for innovative advancements in financial modeling techniques.

In conclusion, our study makes several key contributions to the field of financial DRL. First, we demonstrate the critical importance of temporal window selection in CNN-based models. Second, we show that feature rearrangement can significantly alter the optimal observation period. Finally, we provide a methodological framework for systematically exploring these parameters in future research. These insights open new avenues for enhancing the accuracy and robustness of AI-driven financial analysis tools.